\pdfoutput=1

\documentclass[dvipsnames,format=sigconf]{acmart}

\usepackage{hyperref}
\hypersetup{
	pdftitle={Assignment},
	colorlinks=true,
	bookmarksnumbered=true,
	bookmarksopen=true
}

\usepackage{pifont}
\newcommand{\xmark}{\ding{53}}

\usepackage{listings}
\usepackage{xcolor}
\definecolor{code-green}{rgb}{0,0.5,0}
\definecolor{code-purple}{HTML}{7e57c2}
\definecolor{code-pink}{HTML}{e040fb}
\definecolor{code-light-grey}{HTML}{f6f6f6}
\definecolor{code-dark-grey}{HTML}{606060}
\lstdefinelanguage{ribs}{
    morekeywords={GridArchive, CVTArchive, SlidingBoundariesArchive, EvolutionStrategyEmitter, IsoLineEmitter, GaussianEmitter, Scheduler, BanditScheduler},
    sensitive=true,
    morecomment=[l]{\#},
    morestring=[b]",
}
\usepackage[ruled,commentsnumbered,linesnumbered,noend]{algorithm2e} %
\DontPrintSemicolon
\SetArgSty{textrm} %

\usepackage{subcaption}
\usepackage{multicol}

\AtBeginDocument{%
  \providecommand\BibTeX{{%
    \normalfont B\kern-0.5em{\scshape i\kern-0.25em b}\kern-0.8em\TeX}}}

\copyrightyear{2023}
\acmYear{2023}
\setcopyright{rightsretained}
\acmConference[GECCO '23]{Genetic and Evolutionary Computation Conference}{July 15--19, 2023}{Lisbon, Portugal}
\acmBooktitle{Genetic and Evolutionary Computation Conference (GECCO '23), July 15--19, 2023, Lisbon, Portugal}\acmDOI{10.1145/3583131.3590374}
\acmISBN{979-8-4007-0119-1/23/07}

\newcommand\sferes{Sferes$_{v2}$}

\newcommand\mapelites{\mbox{MAP}-\mbox{Elites}}

\usepackage{todonotes}
\setuptodonotes{size=\footnotesize}

\newcommand{\xxnote}[3]{}
\ifx\hidenotes\undefined%
  \renewcommand{\xxnote}[3]{\color{#2}{(#1: #3)}}
\fi

\definecolor{lightgrey}{rgb}{0.9, 0.9, 0.9}

\newcommand{\sref}[1]{Sec.~\ref{#1}}

\newcommand{\fref}[1]{Fig.~\ref{#1}}
\newcommand{\tref}[1]{Table~\ref{#1}}
\newcommand{\aref}[1]{Algorithm~\ref{#1}}

\def\vnabla{{\bm{\nabla}}}

\usepackage{booktabs}
\usepackage{multirow}
\usepackage{array}
\newcolumntype{L}[1]
  {>{\raggedright\let\newline\\\arraybackslash\hspace{0pt}}m{#1}}
\newcolumntype{C}[1]
  {>{\centering\let\newline\\\arraybackslash\hspace{0pt}}m{#1}}
\newcolumntype{R}[1]
  {>{\raggedleft\let\newline\\\arraybackslash\hspace{0pt}}m{#1}}

\usepackage{amsmath,amsfonts,bm}

\def\eqref#1{equation~\ref{#1}}

\def\1{\bm{1}}

\def\vtheta{{\bm{\theta}}}

\def\vm{{\bm{m}}}

\def\vs{{\bm{s}}}

\DeclareMathAlphabet{\mathsfit}{\encodingdefault}{\sfdefault}{m}{sl}
\SetMathAlphabet{\mathsfit}{bold}{\encodingdefault}{\sfdefault}{bx}{n}

\def\gD{{\mathcal{D}}}

\newcommand{\R}{\mathbb{R}}

\begin{document}

\title[pyribs: A Bare-Bones Python Library for Quality Diversity Optimization]{pyribs: A Bare-Bones Python Library for ~\\ Quality Diversity Optimization}

\author{Bryon Tjanaka}
\orcid{0000-0002-9602-5039}
\email{tjanaka@usc.edu}
\affiliation{%
  \institution{University of Southern California}
  \city{Los Angeles}
  \state{California}
  \country{USA}
}

\author{Matthew C. Fontaine}
\orcid{0000-0002-9354-196X}
\email{mfontain@usc.edu}
\affiliation{%
  \institution{University of Southern California}
  \city{Los Angeles}
  \state{California}
  \country{USA}
}

\author{David H. Lee}
\orcid{0009-0003-1927-5039}
\email{dhlee@usc.edu}
\affiliation{%
  \institution{University of Southern California}
  \city{Los Angeles}
  \state{California}
  \country{USA}
}

\author{Yulun Zhang}
\email{yulunzhang@cmu.edu}
\orcid{0000-0003-3199-8697}
\affiliation{%
  \institution{Carnegie Mellon University}
  \city{Pittsburgh}
  \state{Pennsylvania}
  \country{USA}
}

\author{Nivedit Reddy Balam}
\email{nbalam@usc.edu}
\orcid{0000-0002-1763-4734}
\affiliation{%
  \institution{University of Southern California}
  \city{Los Angeles}
  \state{California}
  \country{USA}
}

\author{Nathaniel Dennler}
\email{dennler@usc.edu}
\orcid{0000-0002-7540-1402}
\affiliation{%
  \institution{University of Southern California}
  \city{Los Angeles}
  \state{California}
  \country{USA}
}

\author{Sujay S. Garlanka}
\email{garlanka@usc.edu}
\orcid{0009-0003-5523-0030}
\affiliation{%
  \institution{University of Southern California}
  \city{Los Angeles}
  \state{California}
  \country{USA}
}

\author{Nikitas Dimitri Klapsis}
\email{nklapsis@usc.edu}
\orcid{0009-0002-2851-0435}
\affiliation{%
  \institution{University of Southern California}
  \city{Los Angeles}
  \state{California}
  \country{USA}
}

\author{Stefanos Nikolaidis}
\email{stefanosnikolaidis@gmail.com}
\orcid{0000-0002-8617-3871}
\affiliation{%
  \institution{University of Southern California}
  \city{Los Angeles}
  \state{California}
  \country{USA}
}

\renewcommand{\shortauthors}{Tjanaka et al.}

\begin{abstract}

Recent years have seen a rise in the popularity of quality diversity (QD) optimization, a branch of optimization that seeks to find a collection of diverse, high-performing solutions to a given problem. To grow further, we believe the QD community faces two challenges: developing a framework to represent the field's growing array of algorithms, and implementing that framework in software that supports a range of researchers and practitioners. To address these challenges, we have developed pyribs, a library built on a highly modular conceptual QD framework. By replacing components in the conceptual framework, and hence in pyribs, users can compose algorithms from across the QD literature; equally important, they can identify unexplored algorithm variations. Furthermore, pyribs makes this framework simple, flexible, and accessible, with a user-friendly API supported by extensive documentation and tutorials. This paper overviews the creation of pyribs, focusing on the conceptual framework that it implements and the design principles that have guided the library's development. Pyribs is available at \url{https://pyribs.org}

\end{abstract}



\maketitle

\begin{figure}[htbp]
\centering
\includegraphics[width=\linewidth]{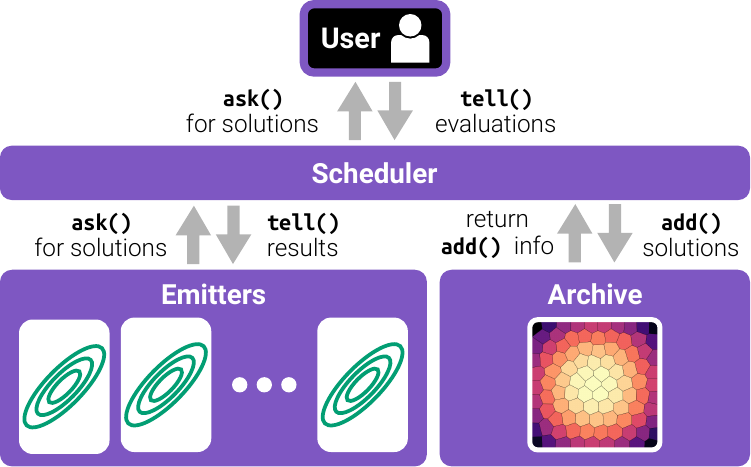}
\Description{Described in caption.}
\caption{Pyribs implements the RIBS framework for QD optimization. The user first \texttt{ask()}'s for solutions from a \emph{scheduler}. The scheduler selects \emph{emitters} to \texttt{ask()} for solutions and returns the solutions to the user. After evaluating the solutions, the user \texttt{tell()}'s the results to the scheduler. The scheduler \texttt{add()}'s the solutions to the \emph{archive} and receives information that it \texttt{tell()}'s to the emitters, enabling the emitters to update their internal search state.}
\label{fig:ribs}
\end{figure}

\begin{table*}

\caption{By selecting different components in the RIBS framework, we can compose a variety of recent algorithms from the QD literature and test them in pyribs. Furthermore, we can identify combinations of components which may lead to new algorithms. Refer to \sref{sec:implement} for more details on the archives, emitters, and schedulers shown here.}
\label{table:components}

{
  \footnotesize
\begin{tabular}{lcc C{0.45in} cccc C{0.45in} C{0.54in} cc}
{} &    \multicolumn{4}{c}{Archive} & \multicolumn{5}{c}{Emitters}    & \multicolumn{2}{c}{Scheduler} \\
\cmidrule(r){2-5}
\cmidrule(r){6-10}
\cmidrule(r){11-12}
{} & Grid & CVT & Sliding Boundaries & Unstructured & Gaussian & Iso+LineDD & CMA-ES & Genetic Algorithm & Gradient Arborescence & Basic & Bandit \\
\midrule
MAP-Elites~\cite{mouret2015map}                 &  \xmark &&&& \xmark &&&&& \xmark \\
CVT-MAP-Elites~\cite{vassiliades2018cvt}        &  & \xmark &&& \xmark &&&&& \xmark \\
 Iso+LineDD MAP-Elites~\cite{vassiliades2018iso} &  & \xmark &&&& \xmark &&&& \xmark \\
MESB~\cite{mesb} & && \xmark && \xmark &&&&& \xmark \\
NSLC~\cite{lehman2011nslc}                      &  &&& \xmark &&&& \xmark && \xmark \\
CMA-ME~\cite{fontaine2020covariance} & \xmark &&&&&& \xmark &&& \xmark \\
CMA-MAE~\cite{fontaine2022mae} & \xmark &&&&&& \xmark &&& \xmark \\
ME-MAP-Elites~\cite{cully2021multi} & \xmark &&&&& \xmark & \xmark &&&& \xmark \\
CMA-MEGA~\cite{fontaine2021dqd} & \xmark &&&&&&&& \xmark & \xmark \\
CMA-MAEGA~\cite{fontaine2022mae} & \xmark &&&&&&&& \xmark & \xmark \\
\end{tabular}
}

\end{table*}

\section{Introduction} %

Many research problems decompose into highly contextual components that prevent one solution from working well across all possible situations.
In such cases, developing a set of solutions rather than a single solution enables researchers to account for a range of contexts.
For instance, a roboticist may develop diverse walking gaits so that their robot can adapt to different morphological considerations~\cite{cully2015}, while a video game designer may generate multiple video game levels so that players can experience various levels of difficulty~\cite{fontaine2021lsi,earle2022}, and a chemist may create multiple viable drug candidates which exhibit unique properties~\cite{me_molecules}.

Quality diversity (QD) optimization~\cite{chatzilygeroudis2021} addresses such problems by searching for collections of diverse, high-performing solutions. Originating in neuroevolution with Novelty Search~\cite{lehman2011ns,lehman2011nslc} and MAP-Elites~\cite{mouret2015map}, QD has grown to become a general-purpose optimization paradigm with applications in a number of areas. As of writing, there are at least 167 papers on the topic~\cite{qdwebsite}, spanning areas as diverse as reinforcement learning~\cite{nilsson2021pga,colas2020scaling,dqdrl,qdpg,tjanaka2022training,conti2018ns}, robot manipulation~\cite{egad,morel2022ns}, human-robot interaction~\cite{fontaine2021quality,fontaine2022evaluating,overcooked}, video game level generation~\cite{fontaine2021lsi,earle2022}, agent testing~\cite{dsage}, generative modeling~\cite{fontaine2021dqd}, urban planning~\cite{archelites}, design~\cite{sail}, internet congestion control~\cite{congestioncontrol}, and drug discovery~\cite{me_molecules}. QD has also moved outside of publications and into more popular forms of media like blog posts~\cite{cabbagecat, game2019blog, frans2021blog, mohamed2021blog, flageat2022blog, allard2021blog, fcmaes2022} and conference tutorials~\cite{qdgecco2020,qdgecco2021,qdgecco2022,qdicml2019}.

To grow further, we believe the QD community must overcome two challenges. The first challenge is to develop a conceptual framework capable of implementing the wide and growing range of QD algorithms.
Many QD algorithms contain interchangeable components, and a unified framework allows for mixing several state-of-the-art components into new algorithms as the field advances.
To this end, previous work has proposed the Unifying Modular Framework (UMF)~\cite{cully2017unifying} to connect the two main families of QD algorithms, Novelty Search and MAP-Elites.
However, UMF was primarily designed for QD algorithms based on genetic operators~\cite{ec}, which limits its applicability to recently developed QD algorithms that have a strong optimization component, such as algorithms which incorporate Evolution Strategies (ES)~\cite{fontaine2020covariance,conti2018ns,tjanaka2022training,dqdrl,colas2020scaling}, gradient ascent~\cite{fontaine2021dqd,fontaine2022mae}, or Bayesian Optimization~\cite{bopelites}.

The second challenge is to implement this framework in software which can support a wide range of users, ranging from beginners entering the field to experienced researchers seeking to develop new algorithms.
Historically, the conception of flexible, well-documented software libraries has been quintessential to the blooming of popular research areas.
For instance, PyTorch~\cite{pytorch} and TensorFlow~\cite{tensorflow} have catalyzed the development and deployment of countless deep learning algorithms in academia and industry~\cite{he2019mlframeworks}, and pycma~\cite{pycma} has popularized the Covariance Matrix Adaptation Evolution Strategy (CMA-ES) as one of the standard tools of evolutionary computation. 
Such libraries have profound effects on their respective fields because not only do they provide powerful features concealed with an expressive, user-friendly application programming interface (API), but they also make these features accessible through comprehensive documentation and tutorials, enabling new practitioners to incorporate the latest algorithms into their projects.
Thus far, the QD community has introduced a number of its own libraries. While these libraries have successfully spurred research, they are targeted towards researchers \emph{within} the QD community, offering them high performance~\cite{mouret2010sferes,lim2022qdax}, reference implementations~\cite{pymap_elites}, or a rich end-to-end experience~\cite{qdpy}.

To address these challenges, we have developed the pyribs library, which implements a conceptual framework that we call RIBS.\footnote{The name ``RIBS'' stems from the title of Fontaine et al.~\cite{fontaine2020covariance}, ``Covariance Matrix Adaptation for the \textbf{R}apid \textbf{I}llumination of \textbf{B}ehavior \textbf{S}pace,'' which introduced the concepts of emitters and schedulers. The name ``pyribs'' is thus a combination of ``Python'' and ``RIBS.'' The proper spelling of pyribs is all-lowercase, similar to pycma~\cite{pycma}, except at the beginning of sentences, when it is capitalized as Pyribs.} 
As shown in \fref{fig:ribs}, a QD algorithm in RIBS is comprised of three components: (1) an \textit{archive} to store solutions generated by the QD algorithm, (2) one or more \textit{emitters} to generate new solutions, and (3) a \textit{scheduler} to manage the interaction of the archive and emitters.

RIBS is highly \textit{modular}: As \tref{table:components} shows, \textit{many existing QD algorithms can be composed by replacing individual components of the framework}. The table also highlights unexplored gaps that could be filled by combining different components, indicating potentially promising areas for future research. Yet, the modular design does not sacrifice simplicity, a key feature in attracting new practitioners.

Moreover, the software implementation of RIBS in pyribs enables seamlessly translating these compositions into code for experimentation and engineering. We achieve this functionality by constructing the library around the following design principles: 
\begin{description}
    \item[Simple:] Centered \textit{only} on components that are absolutely necessary to run a QD algorithm, allowing users to combine the framework with other software frameworks.
    \item[Flexible:] Capable of representing a wide range of current and future QD algorithms, allowing users to easily create or modify components.
    \item[Accessible:] Easy to install and learn, particularly for beginners with limited computational resources.
\end{description}

Pyribs offers modular components that can be assembled into a QD algorithm and controlled with an API inspired by pycma's ask-tell interface~\cite{pycma}. It also features extensive documentation, including tutorials (\fref{fig:tutorials}) demonstrating its usage.\ifdefined\myanon\else\footnote{Website: \url{https://pyribs.org} \\
Source Code: \url{https://github.com/icaros-usc/pyribs} \\
Documentation and Tutorials: \url{https://docs.pyribs.org} %
}\fi\ 
Since its inception in 2021, pyribs has grown to support the research of at least a dozen groups across academia and industry worldwide. As of writing, it has been applied to image generation~\cite{fontaine2021dqd,fontaine2022mae}, video game level generation~\cite{earle2022}, environment generation~\cite{dsage}, reinforcement learning~\cite{tjanaka2022training,dqdrl}, hyperparameter optimization~\cite{qd_hyperparam}, architecture design~\cite{tdomino}, and internet congestion control~\cite{congestioncontrol}.

\section{Background} \label{sec:bkgd} %
\subsection{Quality Diversity}

\subsubsection{Focus} The pyribs library focuses on continuous optimization problems over the search space $\R^n$, the same class of problems targeted by the pycma~\cite{pycma} library. By focusing only on continuous optimization, the library becomes less abstract as search vectors become explicitly defined. Yet, continuous optimization contains an expressive class of problems that the QD community cares about.

\subsubsection{Definition}

We define the continuous QD problem. We assume an \textit{objective function} $f : \R^n \to \R$ and $k$ \textit{measure functions}\footnote{Prior work refers to measure function outputs as ``behavior characteristics,'' ``behavior descriptors,'' or ``feature descriptors.'' We use the ``measures'' terminology in pyribs.} $m_i: \R^n \to \R$, represented jointly as $\vm : \R^n \to \R^k$.  We let  $S = \vm(\R^n)$ be the measure space formed by the range of $\vm$.

The \textit{QD objective} is to find, for each $\vs \in S$, a solution $\vtheta \in \R^n$ such that $\vm(\vtheta) = \vs$ and $f(\vtheta)$ is maximized:
\begin{equation}
    \begin{aligned}
    \max \quad & f(\vtheta) \\ %
    \textrm{subject to} \quad & \vm(\vtheta)=\vs \quad & \forall  \vs \in S
    \end{aligned} \label{eq:qd}
\end{equation}
However, since $S$ is continuous, this objective would require infinite memory to solve, so we relax the QD objective to finding an \textit{archive} (i.e., a finite set) of \textit{representative} solutions $\Theta \subseteq \R^n$. %

A special case of the QD problem is the \textit{differentiable quality diversity (DQD)}~\cite{fontaine2021dqd} problem, where the objective and measure functions are first-order differentiable with gradients $\vnabla f$ and $\vnabla \vm$.

\subsubsection{Algorithms} We consider two alternatives of what constitutes a \textit{representative} solution in the QD problem definition (Eq.~\ref{eq:qd}), resulting in two families of algorithms. 

Algorithms based on MAP-Elites~\cite{mouret2015map} tessellate the measure space $S$ into $M$ cells, and $\Theta$ is constrained such that each of its solutions falls into a different cell of the tessellation based on its measure values.  The vanilla MAP-Elites~\cite{mouret2015map} mutates randomly sampled solutions in the archive with a genetic operator; generated solutions are added to the archive if their objective value exceeds that of the solution currently occupying their corresponding archive cell.
Since its inception, MAP-Elites extensions have included new genetic operators, such as the Iso+LineDD operator inspired by crossover~\cite{vassiliades2018iso}, as well as new methods for tessellating the measure space to create the archive.
For example, MAP-Elites with Sliding Boundaries (MESB) adapts the size of grid cells online to reflect the distribution of solutions in measure space~\cite{mesb}, while CVT-MAP-Elites~\cite{vassiliades2018cvt} precomputes a centroidal Voronoi tessellation (CVT)~\cite{cvt} of the measure space that defines the archive cells.

Algorithms based on Novelty Search~\cite{lehman2011ns,lehman2011nslc} maintain an unstructured archive where each solution must be \textit{novel} by being a certain distance away from its nearest neighbors in measure space.
A genetic algorithm then optimizes a population of solutions to achieve further novelty.
While Novelty Search itself is a purely diversity-driven approach, many of its successors are designed for QD; for instance, Novelty Search with Local Competition (NSLC)~\cite{lehman2011nslc} balances between optimizing for the objective and novelty via multi-objective evolutionary algorithms.

QD algorithms have started to incorporate modern optimization algorithms. For example, Covariance Matrix Adaptation MAP-Elites (CMA-ME)~\cite{fontaine2020covariance} directly optimizes for the QD objective with CMA-ES~\cite{hansen2016cmaes}. In QD optimization, it is efficient to search \textit{multiple} regions of the measure space simultaneously, while balancing the exploration of each region. Therefore, CMA-ME introduced the concepts of \textit{emitters} and \textit{schedulers}. Each emitter maintains a separate CMA-ES instance, while the scheduler balances how emitters explore each measure space region.
Emitters and schedulers became core components of the RIBS framework (\sref{sec:framework}).
Subsequent works building on CMA-ME include Covariance Matrix Adaptation MAP-Annealing (CMA-MAE)~\cite{fontaine2022mae}, which adds an \textit{archive learning rate} to the MAP-Elites grid archive.
The learning rate regulates how quickly a non-stationary discount function changes, resulting in a \textit{soft archive} that balances the tradeoff between pure optimization and exploration.
In addition, CMA-MEGA and CMA-MAEGA (CMA-ME / CMA-MAE via a Gradient Arborescence)~\cite{fontaine2021dqd,fontaine2022mae} address DQD problems with similar principles as CMA-ME and CMA-MAE.

Finally, Multi-Emitter MAP-Elites (ME-MAP-Elites)~\cite{cully2021multi} introduced a new scheduler by modifying the method for selecting emitters.
While the scheduler in CMA-ME maintains several CMA-ES emitters and a round-robin emitter scheduler, ME-MAP-Elites maintains an \textit{emitter pool} consisting of emitters from CMA-ME and emitters that apply the Iso+LineDD operator~\cite{vassiliades2018iso}.
Every iteration, the scheduler uses a multi-armed bandit selector from prior work~\cite{gaier2020blackbox} to select emitters which are likely to improve the archive.

\subsection{The Unifying Modular Framework} \label{sec:umf}

The Unifying Modular Framework (UMF)~\cite{cully2017unifying}, an early conceptual QD framework, proposed to unite the components of the two pioneering algorithms in QD optimization: MAP-Elites and NSLC.

In UMF, QD algorithms consist of a \textit{container} --- equivalent to a RIBS archive --- and a \textit{selector}.
On each iteration of a QD algorithm in UMF, the selector generates solutions that are passed through \textit{random variation} (e.g., mutation or crossover), evaluated, and then inserted into the container.
Containers include the MAP-Elites grid and NSLC unstructured archive, and selection mechanisms include choosing solutions uniformly at random from the container, as in vanilla MAP-Elites, or selecting from a population as in NSLC.

UMF unified under one framework the two major families of QD algorithms: MAP-Elites and NSLC. However, UMF was proposed when all QD algorithms were based on genetic algorithms, and the framework is not expressive enough to represent modern QD algorithms based on other optimization methods. Specifically, UMF incorporates a selector, which chooses solutions as inputs to genetic operators. While selectors can retain a population of solutions, they are not suitable for optimization algorithms that require an internal state, e.g., an evolution path in CMA-ES~\cite{hansen2016cmaes} or momentum in Adam~\cite{adam}. Drawing from the architecture of CMA-ME~\cite{fontaine2020covariance}, our proposed RIBS framework incorporates emitters, which were designed to encapsulate any optimization algorithm used to generate solutions.

In addition, UMF is not designed to manage multiple populations simultaneously. However, algorithms like CMA-ME and ME-MAP-Elites require this feature to maintain multiple CMA-ES instances. The RIBS framework overcomes this design limitation by incorporating a scheduler, which manages multiple emitters.

\subsection{Existing QD Libraries} %

Here we review libraries developed by the QD community, including their goals, features, and relation to pyribs.

\subsubsection{\sferes{}}
\sferes{}~\cite{mouret2010sferes} is a C++ framework for evolutionary computation that also supports QD algorithms.
\sferes{} is primarily designed for high performance, leveraging template-based meta-programming to provide an efficient object-oriented interface and offering multi-core parallel execution through Intel TBB and MPI.
While the template-based structure results in significant performance benefits, it limits accessibility for non-expert users.
In comparison, pyribs focuses solely on QD algorithms rather than on general evolutionary computation. It is a Python-based library that emphasizes accessibility over performance (\sref{sec:accessibility}).

\subsubsection{QDpy}

QDpy~\cite{qdpy} is designed to be a feature-rich Python library for QD optimization.
Besides supporting ready-to-go implementations of algorithms such as MAP-Elites and CMA-ME, QDpy provides building blocks that can be assembled into new algorithms.
To run a QD algorithm, a QDpy user instantiates a \textit{container} (i.e., an archive) and passes it to an \textit{algorithm} object.
The user then defines an evaluation function and passes the function to the QDpy system to optimize.
QDpy also provides logging and plotting utilities and tools to run the evaluation function on distributed computation.

QDpy's flexibility is limited by the requirement that users pass in an evaluation function. While passing in this function allows users to leverage QDpy's various utilities, this requirement also makes it difficult for users to integrate their own utilities.
In contrast, pyribs provides an ask-tell interface where users handle evaluations on their own (\sref{sec:evaluations}).
Essentially, pyribs focuses on components necessary for running QD algorithms, allowing users to integrate tools and frameworks with which they are already familiar.

\subsubsection{pymap\_elites}
pymap\_elites~\cite{pymap_elites} provides customizable reference implementations of MAP-Elites and its variants CVT-MAP-Elites~\cite{vassiliades2018cvt}, MAP-Elites with the Iso+LineDD operator~\cite{vassiliades2018iso}, and Multi-task MAP-Elites \cite{mouret2020multi-task}. Unlike pymap\_elites, pyribs offers a larger selection of algorithms under one framework.

\subsubsection{QDax}
QDax~\cite{lim2022qdax} is a recent library that was developed after the initial release of pyribs. The library focuses on efficient QD, reinforcement learning (RL), and evolutionary algorithm implementations for hardware accelerators such as GPUs and TPUs, taking advantage of the parallel nature of these methods.
QDax specializes in reinforcement learning and robotics domains, where evaluation remains an expensive bottleneck.
Many experiments that took hours or days on a CPU cluster take only minutes with GPU acceleration in QDax.
To leverage accelerators in both function evaluation and algorithm implementation, QDax builds on the JAX library~\cite{jax2018github} and provides a JAX-based API.

While pyribs incorporates batch operations like those found in QDax to ensure a reasonable level of performance (\sref{sec:batch}), pyribs only runs single-threaded on a single CPU (\sref{sec:cpu}). In addition, pyribs is not based on specialized libraries, which makes it accessible to a more general audience, such as beginners who have only basic Python knowledge and limited computational resources. Finally, while QDax extends beyond QD by providing baseline algorithms from RL and multi-objective optimization, pyribs focuses on general-purpose QD algorithms under the RIBS framework.

\section{The RIBS Framework} \label{sec:framework}
Pyribs implements the conceptual RIBS framework that consists of three core components: (1) an \textit{archive} storing solutions generated by the QD algorithm, (2) \textit{emitters} generating solutions for evaluation, and (3) a \textit{scheduler} managing the interaction of the archive and emitters and providing the primary ask-tell~\cite{pycma} interface to the user.
\aref{alg:ribs} shows the standard execution loop for combining these components.
As we show in \sref{sec:dqd}, this execution loop is flexible and not limited to a single call to the ask-tell interface.

\subsection{Components} \label{sec:components}

\subsubsection{Archive} \label{sec:archives}
The archive is a data structure which stores solutions generated by the QD algorithm, along with any information relevant to solutions, such as objective and measure values.
The primary archive method is \texttt{add()}, which takes in multiple solutions with their objective and measure values, attempts to add them to the collection of solutions, and returns information about the addition.
Examples of such information include ``status'' (whether the solution found a new cell in the archive, improved an existing cell, or was not added at all), ``novelty'' (the average distance in measure space from the solution to its $k$-nearest neighbors in the archive~\cite{lehman2011ns}),  and ``improvement value'' (the difference between the solution's objective value and that of the solution which it replaced~\cite{fontaine2022mae}).
Archives may support additional functionality, such as methods for sampling solutions and retrieving solutions with given measure values.

An important choice in the implementation of \texttt{add()} is the order of inserting solutions. The simplest choice is to insert solutions \textit{sequentially}, i.e., one after another. Pyribs offers sequential addition but defaults to the alternative of inserting all solutions simultaneously as a \textit{batch}.

Batching has the following benefits. First, some metrics depend on the order in which solutions are inserted. For example, if two solutions $\vtheta_a$ and $\vtheta_b$ have similar measures, then $\vtheta_a$ may be inserted with high novelty, while $\vtheta_b$ is subsequently inserted with low novelty because $\vtheta_a$ is already in the archive. Batching overcomes this issue by ``freezing'' the archive, then computing the metrics  of all solutions with respect to the frozen archive. Second, batching enables enhanced performance, as libraries like NumPy (used in pyribs) and JAX (used in QDax) are designed to operate on batches of data.

\subsubsection{Emitters} \label{sec:emitters}

QD algorithms in RIBS instantiate one or more emitters. Emitters are algorithms that generate solutions and adapt to objective, measure, and archive insertion feedback. Emitters in RIBS provide two methods. The \texttt{ask()} method queries the emitter's algorithm for candidate solutions. The \texttt{tell()} method updates the internal algorithm state based on the objective and measure values of the generated solutions and any information gained from adding the solutions to the archive.

One example of a RIBS emitter is the CMA-ES emitter from CMA-ME~\cite{fontaine2020covariance}. Here, calling \texttt{ask()} samples solutions from the Gaussian distribution maintained by CMA-ES, while calling \texttt{tell()} updates the Gaussian distribution and internal CMA-ES parameters~\cite{hansen2016cmaes}.

It is also possible that emitters in RIBS do not require any internal state. For instance, when \texttt{ask()} is called, one variation of \mapelites{} generates new solutions by sampling existing archive solutions and perturbing them with fixed-variance Gaussian noise. Since there are no parameters to update for this Gaussian noise mutation, the \texttt{tell()} method does not perform any operation.

\subsubsection{Scheduler}
The scheduler performs two roles in the RIBS framework. First, the scheduler facilitates the interaction between the archive and the population of emitters. The scheduler adds solutions generated by emitters to the archive and passes the results of evaluation and archive insertion to the emitters. Second, schedulers select which emitters generate new solutions on each iteration of the algorithm. Schedulers make decisions on active emitters based on how well each emitter performs in previous iterations.

Schedulers implement an ask-tell interface as shown in \aref{alg:ribs}. When \texttt{ask()} is called (\autoref{line:schedulerask}), the scheduler selects one or more emitters and calls each emitter's \texttt{ask()} method to generate solutions. When \texttt{tell()} is called (\autoref{line:schedulertell}), the scheduler takes in the objective and measure function evaluations of these solutions and \texttt{add()}'s the solutions to the archive. Then, the scheduler passes the solutions, evaluations, and archive addition information to the emitters via each emitter's \texttt{tell()} method.

In the original emitter implementation~\cite{fontaine2020covariance}, emitters directly called \texttt{add()} to insert solutions into the archive. However, allowing emitters to modify the archive meant that feedback from \texttt{add()} depended on the order in which emitters were called, similar to adding solutions sequentially in archives as discussed in \sref{sec:archives}. Now, although the emitters may read data from the archive (e.g., when sampling solutions), only the scheduler calls \texttt{add()} and passes the returned information to the emitters through their \texttt{tell()} method.

Ultimately, the scheduler provides the primary user interface in the RIBS framework. As shown in \aref{alg:ribs}, users directly call \texttt{ask()}, evaluate solutions, and pass the results to \texttt{tell()}.

\begin{algorithm}[t]

\caption{Standard Execution Loop in RIBS}
\label{alg:ribs}

\SetKwProg{QDAlgo}{QD Algorithm}{:}{}
\QDAlgo{$(n_e, n_{it})$}{
  \KwIn{Number of emitters $n_e$, number of iterations $n_{it}$, parameters for $Archive$, $Emitters$, and $Scheduler$}
  \KwResult{Generates solutions to optimize the QD objective, stored in an $Archive$}

  $Archive \gets$ init\_archive$()$ \;
  $[Emitter_1 .. Emitter_{n_e}] \gets$ init\_emitters$(Archive)$ \;
  $Scheduler \gets$ init\_scheduler$(Archive,$ \\
  \hspace{1.43in} $[Emitter_1 .. Emitter_{n_e}])$ \;
  \For{$itr \gets 1..n_{it}$}{ \label{line:loop}
    $L \gets Scheduler.\text{ask}()$ \;
    $User$ computes $Evals = [f(\vtheta), \vm(\vtheta) \text{ for } \vtheta \text{ in } L]$ \;
    $Scheduler.\text{tell}(Evals)$ \;
  }

  \Return $Archive$ \;
}

\SetKwProg{SchedulerAsk}{Scheduler.ask}{:}{}
\SchedulerAsk{$()$}{ \label{line:schedulerask}
  \KwResult{Returns a list of solutions $L$ generated by the emitters.}
  $L \gets []$ \tcp*{Empty list}
  \For{$i \gets 1..n_e$}{
    \If{$Emitter_i$ should generate solutions}{
      $L_i \gets Emitter_i.\text{ask}()$ \;
      $L \gets LL_i$ \tcp*{Concatenate $L_i$ to $L$}
    }
  }
  \Return $L$\;
}

\SetKwProg{SchedulerTell}{Scheduler.tell}{:}{}
\SchedulerTell{$(Evals)$}{ \label{line:schedulertell}
  \KwIn{Objective and measure function evaluations of the list of solutions $L$.}
  \KwResult{Inserts solutions into $Archive$ and updates $Emitters$.}
  $add\_info \gets Archive.\text{add}(L, Evals)$ \;
  \For{$i \gets 1..n_e$}{
    \If{$Emitter_i$ generated solutions}{
      Retrieve solutions $L_i$ generated by $Emitter_i$ \;
      Retrieve $Evals_i$ corresponding to $L_i$ \;
      Retrieve $add\_info_i$ corresponding to $L_i$ \;
      $Emitter_i.\text{tell}(L_i, Evals_i, add\_info_i)$ \;
    }
  }
}

\end{algorithm}

\subsection{Composing Algorithms in RIBS} \label{sec:implement}

\aref{alg:ribs} shows a standard execution loop in RIBS. First, the user configures the core components. Then, in the main loop (\autoref{line:loop}), the user calls the scheduler's ask-tell interface and evaluates solutions in between the calls. Importantly, the RIBS components (archive, emitters, and scheduler) in this loop are interchangeable, and the execution loop can be customized to support new QD algorithms.
We show how replacing components or modifying the execution loop enables RIBS to support a variety of QD algorithms.

\subsubsection{Integrating Different Components} \label{sec:integrating}
First, we consider algorithms which replace components of RIBS without modifying the standard execution loop of \aref{alg:ribs}. \tref{table:components} summarizes the components required for each algorithm. Throughout this section, we italicize the components listed in \tref{table:components} as we introduce them.

We begin with MAP-Elites~\cite{mouret2015map}, which has a \textit{grid archive} that tessellates the measure space into a grid. MAP-Elites incorporates a single emitter that randomly selects solutions from the archive and applies mutations. One kind of mutation is to add Gaussian noise; in this case, we call the emitter the \textit{Gaussian emitter}.
As is common in many versions of MAP-Elites, the Gaussian emitter can also sample directly in the solution space on initial calls to \texttt{ask()}.
Since this emitter has no adaptive components, its \texttt{tell()} method does nothing.
Finally, MAP-Elites has a \textit{basic scheduler} that simply selects this emitter on every iteration.

Replacing components creates different MAP-Elites variants.
Substituting the Gaussian emitter with the \textit{Iso+LineDD emitter}, which applies the Iso+LineDD operation~\cite{vassiliades2018iso}, results in Iso+LineDD MAP-Elites.\footnote{The original Iso+LineDD MAP-Elites~\cite{vassiliades2018iso} uses a CVT archive, but the authors noted that a grid archive would also work with their algorithm.}
We can replace the archive with a \textit{CVT archive} or \textit{sliding boundaries archive} to obtain CVT-MAP-Elites~\cite{vassiliades2018cvt} and MESB~\cite{mesb}.

We can also consider methods based on Novelty Search like NSLC~\cite{lehman2011nslc}. Here, the \textit{unstructured archive} adds solutions if they are far away from their $k$ nearest neighbors in the archive. Meanwhile, the \textit{genetic algorithm emitter} contains a genetic algorithm such as NEAT~\cite{neat} that manages a population of solutions.
In contrast to the Gaussian and Iso+LineDD emitters, the genetic algorithm emitter's \texttt{tell()} method updates its internal population. The scheduler remains the same as in MAP-Elites.

CMA-ME~\cite{fontaine2020covariance} and CMA-MAE~\cite{fontaine2022mae} are more complicated because they require managing multiple instances of CMA-ES in parallel. In this case, we create multiple \textit{CMA-ES emitters}, each with their own CMA-ES instance. Calling \texttt{ask()} on each emitter samples solutions from CMA-ES's multivariate Gaussian distribution, and calling \texttt{tell()} updates the distribution parameters and the internal CMA-ES parameters. We combine these emitters with the grid archive and basic scheduler from MAP-Elites.

Multi-Emitter MAP-Elites (ME-MAP-Elites)~\cite{cully2021multi} provides an example of an algorithm that requires a different scheduler. The default ME-MAP-Elites includes CMA-ES and Iso+LineDD  emitters. Its \textit{bandit scheduler} applies a multi-armed bandit algorithm to select a subset of these emitters based on whether they have previously generated solutions that were inserted into the archive.

\begin{algorithm}[t]

\caption{QD Algorithm with Surrogate Model in RIBS}
\label{alg:ribs_surrogate}

\SetKwProg{QDAlgo}{QD \hspace{-0.006in}Algorithm \hspace{-0.006in}with \hspace{-0.006in}Surrogate \hspace{-0.006in}Model}{:}{}
\QDAlgo{\hspace{-0.015in}$(n_e, n_{inner}, n_{outer})$}{
  \KwIn{Number of emitters $n_e$, inner loop iterations $n_{inner}$, outer loop iterations $n_{outer}$, parameters for $Archive$, $Emitters$, $Scheduler$, and $Model$}
  \KwResult{Generates solutions to optimize the QD objective, stored in an $Archive$}

  $Archive \gets$ init\_archive$()$ \label{line:archveinit} \;
  $Model \gets$ init\_surrogate\_model$()$ \;
  $\gD \gets \{\}$ \tcp*{Dataset of evaluated solutions}

  \For{$itr \gets 1..n_{outer}$}{ \label{line:outerloop}
      \tcp{Construct surrogate archive.}
      $Archive' \gets$ init\_archive$()$ \;
      $[Emitter'_1 .. Emitter'_{n_e}] \gets$ init\_emitters$(Archive')$ \;
      $Scheduler' \hspace{-0.03in} \gets$ \hspace{-0.03in} init\_scheduler$(Archive',$ \\
      \hspace{1.42in} $[Emitter'_1 .. Emitter'_{n_e}])$\;
      \For{$iter \gets 1..n_{inner}$}{ \label{line:innerloop}
        $L \gets Scheduler'.\text{ask}()$ \;
        $Evals' \hspace{-0.01in} \gets [Model.f(\vtheta), Model.\vm(\vtheta) \text{ for } \vtheta \text{ in } L]$\label{line:surrogateeval}\;
        $Scheduler'.\text{tell}(Evals')$ \;
      }
      \tcp{Record true evaluations of solutions.}
      $L \gets$ all solutions in $Archive'$ \;
      $User$ computes $Evals = [f(\vtheta), \vm(\vtheta) \text{ for } \vtheta \text{ in } L]$ \;
      $Archive.\text{add}(L, Evals)$ \label{line:archiveadd} \;
      \tcp{Update model.}
      $\gD \gets \gD \cup (L, Evals)$ \;
      Train $Model$ on data in $\gD$ \label{line:modeltrain} \;
  }

  \Return $Archive$ \;
}

\end{algorithm}

\subsubsection{Modifying the execution loop} \label{sec:dqd}

Besides algorithms that replace components of RIBS, we also consider those that modify the RIBS execution loop. In this regard, CMA-MEGA~\cite{fontaine2021dqd} and CMA-MAEGA~\cite{fontaine2022mae} both require a \textit{gradient arborescence emitter}, which constructs solutions by branching from a solution point based on the objective and measure gradients. This branching requires calling \texttt{ask()} and \texttt{tell()} twice: once to collect the solution point and return its evaluations and gradients, and once to handle the branched solutions. Compared to \aref{alg:ribs}, we add another set of calls to \texttt{ask()} and \texttt{tell()} in the loop on \autoref{line:loop}, with appropriate arguments to handle passing gradients back to \texttt{tell()}.

In addition, a number of recent works~\cite{sail,dsame,dsage,bopelites} integrate surrogate models with QD algorithms in domains where evaluations are expensive. Surrogate-assisted QD algorithms construct an archive based on evaluations predicted by a surrogate model and then select candidate solutions for ground-truth evaluations.

\aref{alg:ribs_surrogate} shows a general layout for such an algorithm. This algorithm maintains a \textit{ground-truth archive} for storing solutions which have been evaluated by the user (\autoref{line:archveinit}).
Then, during an \textit{outer loop} (\autoref{line:outerloop}), it performs three phases. First, it constructs a \textit{surrogate archive} in an \textit{inner loop} (\autoref{line:innerloop}) based on solutions evaluated by the model (\autoref{line:surrogateeval}). Second, the user evaluates the candidate solutions from the surrogate archive, and the evaluated solutions are added into the ground-truth archive (\autoref{line:archiveadd}). Finally, the algorithm trains the model to improve its predictions (\autoref{line:modeltrain}).

\section{Designing pyribs} \label{sec:design}

To realize the RIBS framework, we created the pyribs library and released it in 2021.  The structure of the library closely follows the framework, with subpackages for archives, emitters, and schedulers. We describe the principles that have guided our implementation decisions and notable features that highlight these principles.

\begin{figure*}
\centering
\includegraphics[width=\linewidth]{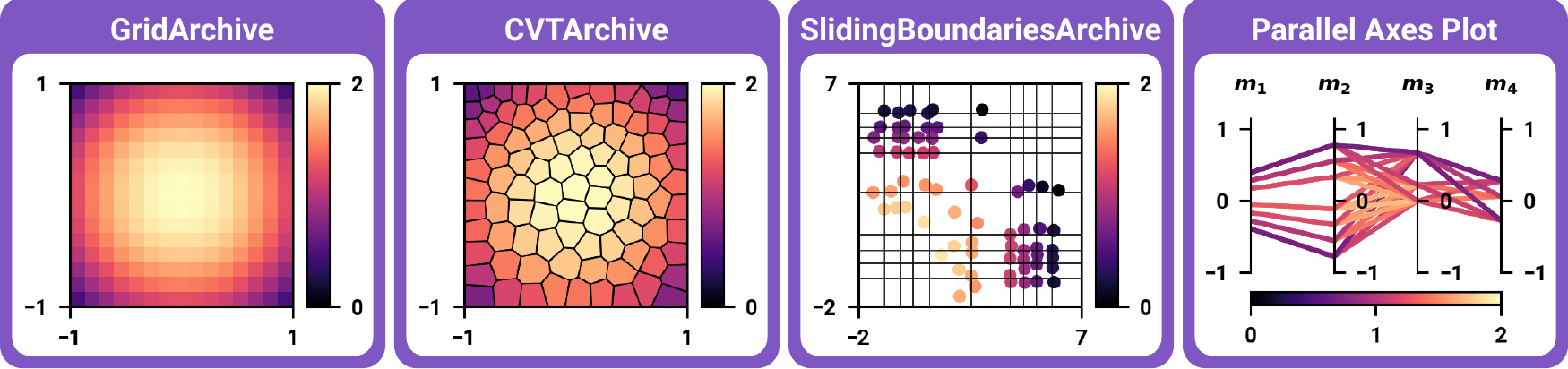}
\Description{Four visualization tools from pyribs. The first is a grid archive heatmap. The second is a CVT archive heatmap. The third is a sliding boundaries archive heatmap. The fourth is a parallel axes plot. These plots are described in the caption.}
\caption{Pyribs visualization tools. We show example 2D heatmaps, where the axes correspond to the measure values, and the color of each archive cell indicates its objective value. In \texttt{SlidingBoundariesArchive}, the points show the locations of solutions in measure space, and the lines show the grid boundaries. We also show a \textit{parallel axes plot} which can visualize an archive of any dimensionality. In this plot, a single solution's measures are plotted as a line connecting the measures $m_1\ldots m_k$, and the line is colored according to the solution's objective value.}
\label{fig:heatmaps}
\end{figure*}

\subsection{Principles} \label{sec:principles}

\subsubsection{Simple}
We designed pyribs to be ``bare-bones'' and maintain only the core components required for a QD algorithm optimizing a continuous search space. The simplicity of the design makes the library easier for new users to adopt, while the focus on continuous optimization problems reduces implementation complexity and makes the defined search space less abstract. 

\subsubsection{Flexible}
Pyribs is also ``bare-bones'' in the sense that the core components of the library --- archives, emitters, and schedulers --- are all exposed to the user. This allows users to easily exchange components of the QD algorithm, and the design provides a foundation to implement future QD algorithms discovered by researchers.

\subsubsection{Accessible} \label{sec:accessibility}

Pyribs is accessible to a wide audience, ranging from beginners to experienced researchers, by having readable source code, being easy to install, and having full documentation defining its usage. Our dependency choices ensure that beginners with limited computational resources or basic hardware can install pyribs and study the tutorials. The library also supports experienced researchers by being amenable to modifications.

\subsection{Implementation Features}

These features demonstrate how our implementation choices align with the design principles of section~\ref{sec:principles}.

\subsubsection{Choice of Python} 
Python offers many desirable features. Beyond being a beginner-friendly language, it has a flourishing ecosystem, with package repositories like the Python Package Index (PyPI)~\cite{pypi} and Anaconda~\cite{anaconda} providing easy access to many useful libraries.
While Python itself is slower than lower-level languages like C++, libraries like NumPy~\cite{numpy} compensate for this limitation by providing access to efficient numerical computation routines.
Python can also integrate with other programming languages through various packages; for instance, PyJNIus~\cite{pyjnius} enables running Python-based QD algorithms~\cite{fontaine2021lsi,dsage} with the Mario AI Framework~\cite{marioframework}, a benchmark implemented in Java.
Furthermore, Python has become the \textit{de facto} language of machine learning, and with the influx of QD applications to machine learning~\cite{dsame,fontaine2021dqd,nilsson2021pga,fontaine2021lsi}, it is important to support QD researchers from that area.
Thus, implementing the RIBS framework in Python and distributing pyribs on PyPI and Anaconda makes pyribs \textit{accessible}, as users can easily install and learn to use the library.

\subsubsection{Focus on continuous optimization} \label{sec:continuous} To maintain \textit{simplicity}, pyribs only supports continuous optimization problems with a fixed number of parameters.
Such problems are ubiquitous in a variety of fields, including machine learning, and continuous fixed-length solutions are readily represented in software as arrays that can be efficiently manipulated by libraries like NumPy.

There are many other solution encodings that a QD library could support.
For instance, discrete solutions (e.g., a list of integers) can be implemented with the same arrays used in pyribs, but recent research in QD has focused on continuous domains~\cite{chatzilygeroudis2021}.
Alternatively, a QD library could support objects such as solutions of variable length, graphs~\cite{me_molecules}, or neural network structure encodings from NEAT~\cite{neat}.
Although the RIBS framework is not limited to any one solution encoding, such structures are often domain-specific, and including them would increase the complexity of pyribs.

\subsubsection{Single CPU} \label{sec:cpu} To be \textit{simple} and \textit{accessible}, pyribs runs single-threaded on a single CPU. Thus, pyribs can run on hardware ranging from laptops to high-performance clusters.
While being single-threaded may limit the performance of pyribs, the runtime in many QD problems is dominated by the user's evaluation of solutions, rather than by the execution of the QD algorithm in pyribs.
Moreover, the need for high-performance algorithm implementations is already fulfilled by libraries like \sferes{}~\cite{mouret2010sferes} and QDax~\cite{lim2022qdax}.
However, if internal algorithm runtime grows to be a bottleneck in QD problems, we could redesign pyribs to optionally parallelize execution by putting each emitter on a separate thread.
Note that while pyribs itself runs on a single CPU, evaluations are left to the user and may be parallelized as described in the next section.

\subsubsection{Evaluations are left to the user} \label{sec:evaluations}
Since evaluations are often the bottleneck in QD, we considered providing utilities for running evaluations of solutions in parallel, as is done in QDpy~\cite{qdpy}. 
We decided against doing so since many evaluation functions require specific dependencies and hardware configurations that are difficult to support in a general-purpose library.
In short, we maintain \textit{simplicity} by leaving evaluations to the user.
Nevertheless, our documentation includes basic examples of how to integrate parallelism into pyribs workflows.
For instance, our tutorial ``Using CMA-ME to Land a Lunar Lander Like a Space Shuttle'' parallelizes evaluations in only two lines of code with Python's \texttt{multiprocessing} module.

\begin{figure*}
\centering
\includegraphics[width=\linewidth]{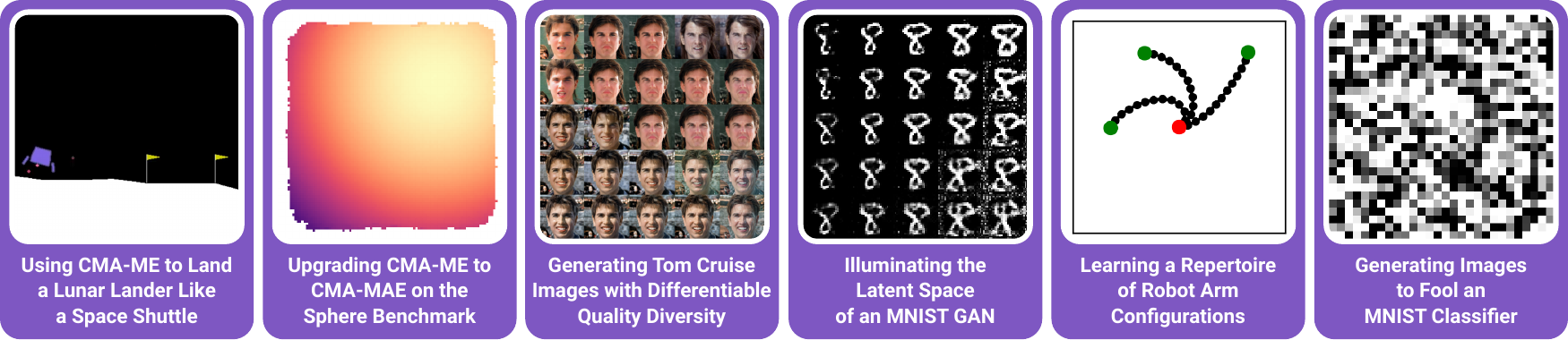}
\Description{Images of six tutorials from pyribs. First is "Using CMA-ME to Land a Lunar Lander Like a Space Shuttle."
Second is "Upgrading CMA-ME to CMA-MAE on the Sphere Benchmark."
Third is "Generating Tom Cruise Images with Differentiable Quality Diversity."
Fourth is "Illuminating the Latent Space of an MNIST GAN."
Fifth is "Learning a Repertoire of Robot Arm Configurations."
Sixth is "Generating Images to Fool an MNIST Classifier."}
\caption{Tutorials enable pyribs users to quickly learn about the library and experiment with problems from the QD literature.}
\label{fig:tutorials}
\end{figure*}

\subsubsection{Batch operations} \label{sec:batch}
The initial version of pyribs implemented many operations sequentially, including the archives' \texttt{add()} methods (\sref{sec:archives}).
To compensate for the reduced performance, we added the just-in-time compiler Numba~\cite{numba} to many of these functions.
However, multiple users indicated that doing so decreased \textit{accessibility} by making the library difficult to modify and debug.
Hence, in the most recent version of pyribs (0.5.0), we have re-implemented these methods as batch operations without Numba.
For instance, instead of adding one solution to the archive at a time, pyribs now leverages NumPy array operations to add a batch of solutions simultaneously.
These operations improve code readability and performance over the sequential implementations while still operating on a single CPU.
For instance, on the 20D sphere linear projection benchmark, the runtime of CMA-ME decreased from 140s with sequential+Numba to 60s with batch implementations.

\subsubsection{One-layer hierarchy} \label{sec:hierarchy}
Ideally, every pyribs component would be implemented in a single file with no dependencies.
Doing so would make the source code highly \textit{accessible}, as a user could easily read the code for a component and modify it, similar to pymap\_elites~\cite{pymap_elites}.
However, since components often share functionality, standalone files would lead to duplicate code and hamper maintenance.
We compromise by introducing a one-layer hierarchy, where archives, emitters, and schedulers all inherit from their respective base classes.
For instance, a user can learn about the implementation of \texttt{GridArchive} by reading the source code for \texttt{ArchiveBase} and \texttt{GridArchive}.
This hierarchy helps make pyribs \textit{flexible}, as users who create new components can inherit from these base classes instead of re-implementing basic functionality.

\subsubsection{Visualization tools} \label{sec:visualization} Since there are no commonly available visualization tools for archives, we added our own to make pyribs \textit{accessible}. As shown in \fref{fig:heatmaps}, pyribs features heatmap visualizations for all its archives, as well as a parallel axes plot method that can visualize an archive of any dimensionality. 

\subsubsection{Documentation and tutorials}
A key feature for increasing \textit{accessibility} in pyribs is its extensive documentation and tutorials. Every pyribs component is documented in detail, and pyribs has an array of tutorials (\fref{fig:tutorials}). These tutorials teach users about the library and introduce them to common QD problems, such as latent space illumination~\cite{fontaine2021lsi}, the arm repertoire benchmark~\cite{cully2017unifying,vassiliades2018iso}, and the sphere linear projection benchmark~\cite{fontaine2020covariance}.

\subsubsection{Industry standard practices} \label{sec:industry}
We draw from industry standard style guides~\cite{googlepystyle} to promote readability and correctness in our source code. We automatically format our code with yapf~\cite{yapf} and check for basic errors with Pylint~\cite{pylint}. 
Furthermore, we implement a comprehensive suite of unit tests. Since it is difficult to test stochastic components like emitters and schedulers, our total code coverage by unit tests\footnote{Code coverage measures the proportion of library code that is executed in tests.} as of version 0.5.0 is 81\%, but on archives, which are nearly deterministic, our coverage is 97\%.
Finally, when implementing new components, we run them on benchmarks such as sphere linear projection~\cite{fontaine2020covariance} to verify that our implementation matches results from prior work. These practices ensure that future changes in pyribs do not affect existing functionality.

\section{Conclusion}

This paper details the design of the conceptual RIBS framework and its implementation in the pyribs library. We show how RIBS supports a wide range of QD algorithms by interchanging core components, and we highlight the design principles of the library --- simplicity, flexibility, and accessibility.

In the long run, our goal is for pyribs to become a library that supports a wide range of users in the QD community.
On one end of the spectrum, we seek to support beginners by making pyribs ever more easy to learn and use.
To this end, we will continue to maintain our documentation and tutorials and incorporate user feedback in our implementation decisions.
Our vision is that pyribs will become an entry point into QD for many researchers.

On the other end of the spectrum, we aim to serve the needs of more experienced researchers by further developing the capabilities of pyribs.
A major avenue in this direction is to expand the collection of pyribs components.
Pyribs currently centers on the MAP-Elites family, but potential additions outside this family include the unstructured archive from Novelty Search, the archive with learned measures from AURORA~\cite{aurora}, and emitters and schedulers from NS-ES~\cite{conti2018ns} and SERENE~\cite{serene}.

Simultaneously, it is important for pyribs to support integrations with other fields.
For example, many recent QD papers combine QD with deep learning~\cite{fontaine2021lsi,fontaine2021dqd,fontaine2022mae,nilsson2021pga,dsage}.
Given the prevalence of GPUs in deep learning, it could thus be helpful to add GPU support to overcome delays incurred by transferring data between CPU and GPU.
However, such advanced features will need to be delicately balanced against our design principles.

Beyond directly supporting practitioners, we believe that the lessons learned from the development of pyribs can inform the design and development of future QD libraries.
We are thus excited about the supporting role that pyribs can play in expanding the QD community and growing QD into a widely adopted discipline.

\begin{acks}
This work was partially supported by the NSF CAREER (\#2145077), NSF NRI (\#2024949), and NSF GRFP (\#DGE-1842487).
We thank the anonymous reviewers, Varun Bhatt, Ya-Chuan Hsu, and Robby Costales for their invaluable feedback.
\end{acks}

\newpage

\bibliographystyle{acm}
\bibliography{references}

\clearpage

\appendix

\section{Logo}

\begin{figure}[htbp]
\centering
\includegraphics[width=0.95\linewidth]{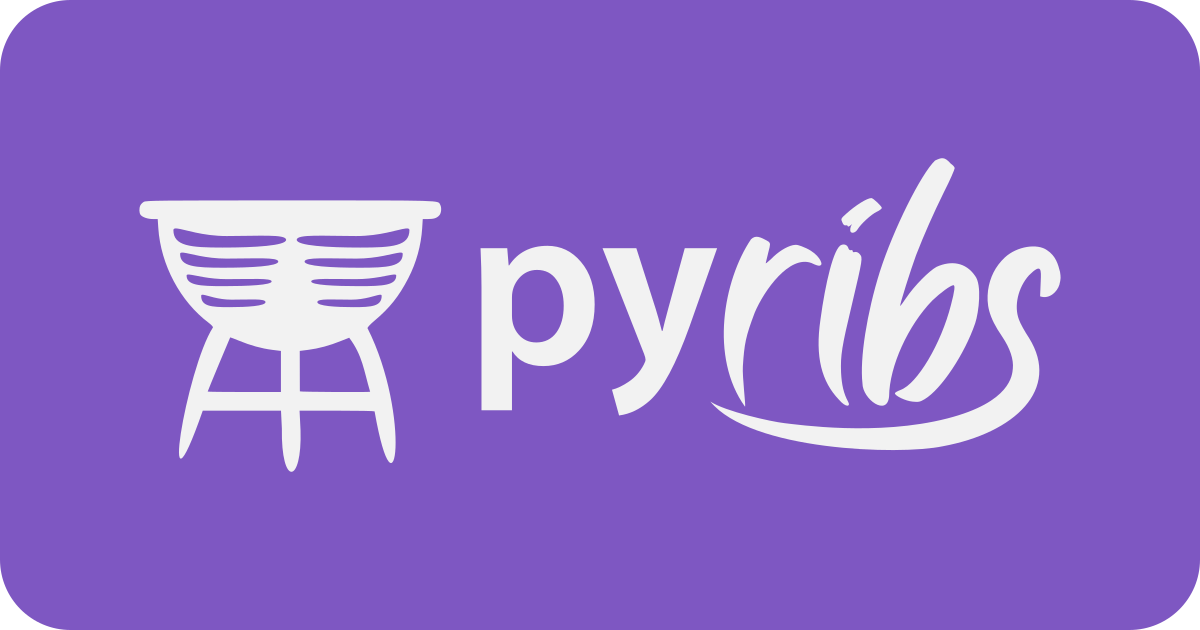}
\caption{The pyribs logo.}
\label{fig:logo}
\end{figure}

\noindent Modern libraries such as PyTorch~\citep{pytorch} have logos to help users recognize and identify the library in talks, tutorials, and other materials. Usually, the design of the logo is somehow connected to the design or name of the library. 

In pyribs, we connect the design of the logo~(\fref{fig:logo}) to the RIBS acronym, as well as to the dual-sense meaning of \textit{bare-bones}. Recall that pyribs is bare-bones in two senses. First, pyribs maintains \textit{simplicity} by only maintaining core components of a QD algorithm. Second, the library is \textit{bare-bones} in the sense that the core components of the library --- archives, emitters, and schedulers --- are all exposed to the user.

We capture a similar spirit in the design of the pyribs logo by depicting ribs in two senses. The logo depicts a smoker, commonly used to cook beef or pork ribs, overlaid with a human rib cage. The smoker depicts the simplicity aspect of \textit{bare-bones}, as pyribs is a useful tool because of its simplicity, while the rib cage captures the exposed nature of core pyribs components, allowing easy modification. The choice of the cooking-themed logo design is also closely connected to the modular nature of pyribs, allowing users to \textit{cook up} new QD algorithms by exchanging components.

The pyribs logo lettering captures another dual-sense nature of pyribs. The RIBS framework is a conceptual framework for QD algorithms. The lettering of ``ribs'' augments a bone typeface to capture the abstract nature of the framework and draw a connection to the \textit{bare-bones} conceptual design of pyribs. Meanwhile, the ``py'' is written in a Python-style typeface to connect with the Python implementation of pyribs.

Overall, we believe that a thoughtful and high-quality logo raises the cachet of pyribs and will help attract both researchers and practitioners to the library.

\newpage

\end{document}